\documentclass{article}
\usepackage{mlspconf,amsmath,graphicx}
\usepackage{epsfig}
\usepackage{graphicx}
\usepackage{amsmath}
\usepackage{amssymb}
\usepackage{mathtools,xparse}
\usepackage{pgfplots}

\usepackage{amsmath,graphicx}
\usepackage{amssymb}
\usepackage{epstopdf}
\usepackage{relsize}
\usepackage{multirow}
\usepackage{float}
\usepackage{lipsum}  
\usepackage{graphicx}
\usepackage{subcaption}

\usepackage{tikz}
\usetikzlibrary{arrows,shapes}
\usetikzlibrary{positioning}
\usepackage[keeplastbox]{flushend}
\usepackage{textcomp}
\usepackage{amsfonts}
\usepackage{textcomp}

\copyrightnotice{To appear in 2018 IEEE International Workshop on Machine Learning for Signal Processing, Sept.\ 17--20, 2018, Aalborg, Denmark}
\toappear{\copyright 2018 IEEE. Only personal use of this material is permitted}

\newlength\figureheight
\newlength\figurewidth
\title{Model-order selection in statistical shape models}

\twoauthors
  {Alma Eguizabal, Peter J. Schreier}
	{Signal and System Theory Group \\
	University of Paderborn, Germany\\
	\{alma.eguizabal, peter.schreier\}@sst.upb.de}
  {David Ram{\'i}rez\thanks{The work of A. Eguizabal and P. J. Schreier was supported in part by the German Research Foundation (DFG) under grant SCHR 1384/3-1. The work of D. Ram{\'\i}rez has been partly supported by Ministerio de Econom{\'\i}a of Spain under projects: OTOSIS (TEC2013-41718-R) and the COMONSENS Network (TEC2015-69648-REDC), by the Ministerio de Econom{\'\i}a of Spain jointly with the European Commission (ERDF) under projects ADVENTURE (TEC2015-69868-C2-1-R) and CAIMAN (TEC2017-86921-C2-2-R), and by the Comunidad de Madrid under project CASI-CAM-CM (S2013/ICE-2845).}}
	{Dept. of Signal Theory and Communications \\
	Universidad Carlos III de Madrid, Spain\\
	david.ramirez@uc3m.es}
\begin{document}
\maketitle

\begin{abstract}
Statistical shape models enhance machine learning algorithms providing prior information about deformation. A Point Distribution Model (PDM) is a popular landmark-based statistical shape model for segmentation. It requires choosing a model order, which determines how much of the variation seen in the training data is accounted for by the PDM. A good choice of the model order depends on the number of training samples and the noise level in the training data set. Yet the most common approach for choosing the model order simply keeps a predetermined percentage of the total shape variation. In this paper, we present a technique for choosing the model order based on information-theoretic criteria, and we show empirical evidence that the model order chosen by this technique provides a good trade-off between over- and underfitting.
\end{abstract}
\section{Introduction}
Statistical shape models provide prior information about the deformation of an object \cite{Luthi13}. A Point Distribution Model (PDM) \cite{Cootes96} contains statistical information of a collection of shape landmarks and its variability, represented by an affine space of eigenvectors obtained by Principal Component Analysis (PCA). Popular segmentation techniques that make use of PDMs are Active Shapes Models  \cite{Santiago15} and Constrained Local Models \cite{Lindner13}. We refer to the number of principal components kept in the PDM as its model order. If the order is too large, the model may not be specific enough (overfitting); if the order is too small, new observations of the same shape may not be accurately represented with the model (underfitting). A good model order provides the right trade-off between overfitting and underfitting. 

The most common way of choosing the model order of a PDM is to keep the eigenvectors that account for a given percentage of variance (typically 90-98\% \cite{Lindner13}). Many landmark-based shapes, such as anatomical shapes in medical image analysis, are high-dimensional, and often only few observations are available. These observations may also contain noise artifacts. The heuristic approach of choosing the model order based on a kept fraction of total variance may therefore be suboptimal. The best model order varies significantly depending on the number of samples and noise level of the training data set. Our motivation in this paper is to design a model-order selection rule that has a theoretical justification and leads to a statistical shape model with good representation ability. We consider therefore information theory, which has successfully been used before to enhance registration and detection algorithms \cite{Viola97}\cite{Zois16}, as well as to place landmarks automatically in statistical shape models \cite{Davies02}.

This model-order selection problem has been addressed before in statistical shape model design. In \cite{Mei08} the authors suggested a t-test of bootstrap stability of the PCA modes of the PDM, and they validated the strategy on simulated anatomical shapes with white noise. The authors in \cite{Nadakuditi08} proposed a strategy based on an information-theoretic criterion for small sample support in a more generic array-processing context, also assuming white noise. These techniques may fail if the noise is not white. In \cite{Aouada04}, also in an array-processing context, the authors considered nonuniform noise.

In PDM design, there is no obvious model for the noise. Therefore, techniques that consider a specific noise structure may not work well. In order to address this, we propose a new strategy, based on information-theoretic criteria, that assumes a more generic colored-noise model. Our strategy is specifically designed to determine the model order in a PDM, although it may also be applied to other model-order selection problems with colored noise. We interpret the PDM as a multivariate regression, where the model order is determined considering the statistical properties of the regression residuals. We also perform a comparative study, with simulated and real shapes, where we prove the good performance of our strategy, as well as the importance of an accurate model order in PDMs. 

The remaining of the paper is organized as follows: Section 2 explains in detail the problem formulation, and how the PDM is interpreted as a regression problem. Section 3 presents the proposed solution based on information theory. Section 4 shows a comparative and validation study with simulated shapes where the model order is known, and shapes from real databases where the model order is unknown. We summarize the conclusions in Section 5.

\section{Problem formulation}
\label{ssec:Form}
A PDM of an object of interest models the variability of its shape \cite{Cootes96}.
 Let the random vector $\mathbf{x} \in \mathbb{R}^N $ model the $x$- and $y$-coordinates of $\frac{N}{2}$ landmarks of a shape after a Procrustes aligment \cite{BOOKSTEIN97}. With the mean $\mathbb{E}[\mathbf{x}] = \boldsymbol{\mu}_\mathbf{x}$, we define the zero-mean shape vector as $\mathbf{y} = \mathbf{x} -\boldsymbol{\mu}_\mathbf{x}$.
 
The $N$ variables in $\mathbf{y}$ are highly correlated, and most of the information on shape variability is contained in a lower-dimensional subspace. Let us define the covariance matrix of the shape vector, $\mathbf{R}_{yy} =  \mathbb{E}[\mathbf{y}\mathbf{y}^T]$, and consider its eigenvalue decomposition, i.e., $\mathbf{R}_{yy} = \mathbf{P}\boldsymbol{\Lambda}\mathbf{P}^T$. 
 We perform dimensionality reduction for $\mathbf{y}$, assuming the following linear model:
\begin{equation} \label{yP}
\mathbf{y} = \mathbf{ P}_t\mathbf{ b}_t + \mathbf{\boldsymbol{\epsilon}},
\end{equation} 
where $\mathbf{b}_t\in \mathbb{R}^t$ is a vector of $t<N$ parameters, and $\mathbf{P}_t = [\mathbf{p}_1, \mathbf{p}_2, \ldots, \mathbf{p}_t] \in \mathbb{R}^{N\times{t}}$ are the $t$ eigenvectors of $\mathbf{R}_{yy}$ corresponding to the $t$ largest eigenvalues. In order to preserve shape plausibility, $\mathbf{b}_t$ is restricted to a set $\mathbb{B}(\boldsymbol{\lambda}_t )$, where $\boldsymbol{\lambda}_t$ is a $t$-dimensional vector containing the $t$ largest eigenvalues of $\mathbf{R}_{yy}$. The vector $\boldsymbol{\epsilon}$ accounts for the variability that we do not represent with the model. We consider $\boldsymbol{\epsilon}$ to be a zero-mean Gaussian noise vector, with arbitrary covariance matrix $\boldsymbol{\Sigma}$, i.e., colored noise. The noise may be due to imprecise training landmarks, presence of non-representative variability in the training set, insufficient samples, or quantization errors. Thus, the noise at different landmarks may be correlated and may have different variances. Furthermore, the prior Procrustes alignment typically introduces color in the noise.  

The problem is to determine the value of the model order $t$. Following the definition of best shape model provided in \cite{Davies02}, we consider an information-theoretic approach to selecting $t$, which provides a trade-off between compactness (size of $t$), specificity (overfitting) and generalization ability (underfitting). 

\section{Proposed Solution}

Let us assume we obtain $M$ observations of the vector $\mathbf{x}$, that is, $M$ shape training samples ${\bf x}^{(m)}$, $m = 1,...,M$, after a Procrustes alignment. Our solution divides the problem into two steps. First, we obtain estimates for the model parameters $\mathbf{P}$, $\boldsymbol{\Lambda}$, and $\boldsymbol{\mu}_{\mathbf{x}}$. Then, we estimate the model order $t$. For this, we split the observed data into two subsets of sizes $M_1$ and $M_2$, denoted by $\mathbf{X}_1 = \big[ \mathbf{x}^{(1)}, \dots, \mathbf{x}^{(M_1)} \big] \in \mathbb{R}^{N\times M_1}$ and $\mathbf{X}_2 = \big[ \mathbf{x}^{(M_1+1)}, \dots, \mathbf{x}^{(M_1+M_2)} \big]\in \mathbb{R}^{N\times M_2}$. In our implementation, we choose $M_1 = M_2$.

Within the first set $\mathbf{X}_1$, we compute the sample mean  $\hat{\boldsymbol{\mu}}_\mathbf{x} = \frac{1}{M_1}\sum_{m=1}^{M_1}\mathbf{x}^{(m)}$ as well as the sample covariance matrix $\hat{\mathbf{R}}_{yy} = \frac{1}{M_1}(\mathbf{X}_1\mathbf{X}_1^T-\hat{\boldsymbol{\mu}}_\mathbf{x}\hat{\boldsymbol{\mu}}_\mathbf{x}^T)$ and its eigenvalue decomposition $\hat{\mathbf{R}}_{yy} = \hat{\mathbf{P}}\hat{\boldsymbol{\Lambda}}\hat{\mathbf{P}}^T$.

Let us define the matrix $\bf Y$ that contains the entries of ${\bf X}_2$ with mean removed, and consider the following multivariate linear regression:
\begin{equation} \label{eq5}
\mathbf{Y} = \hat{\mathbf{P}}_t\mathbf{B}_t + \mathbf{E},
\end{equation}
 where the columns in $\mathbf{B}_t\in \mathbb{R}^{t\times M_2}$ are parameter vectors $\mathbf{b}_t$ as defined in \eqref{yP} corresponding to each of the $M_2$ vectors in set $\mathbf{X}_2$; we model the regression noise in matrix $\mathbf{E} = \big[\boldsymbol{\epsilon}^{(1)},\dots,\boldsymbol{\epsilon}^{(M_2)}\big]$, whose columns are considered independently and identically distributed (i.i.d.) observations of the error vector in \eqref{yP}.  We propose an information-theoretic formulation for model selection in multivariate linear regressions, similarly as described in \cite{Bedrick94}. The model order $t^*$ is chosen as
\begin{equation} \label{eq3}
t^*=\arg\min_{t}( \underbrace{-\log p(\mathbf{Y}|\hat{\mathbf{B}}_t,\hat{\boldsymbol{\Sigma}}_t)}_\text{likelihood term} + \underbrace{\eta(t)}_\text{penalty}),
\end{equation}
where $\hat{\mathbf{B}}_t$ and $\hat{\boldsymbol{\Sigma}}_t$ are the Maximum Likelihood (ML) estimates of the model parameters for model order $t$, and $\eta(t)$ is a penalty term that depends on the selected information criterion (Akaike, Bayesian, etc.) \cite{Stoica04}. We model ${\mathbf{B}}_t$ as an unknown deterministic parameter. Thus, the only random quantity in the regression is the residual noise matrix $\mathbf{E} = \mathbf{Y}-\hat{\mathbf{P}}_t\mathbf{B}_t$. Consequently, the log-likelihood expression in \eqref{eq3} can be written as \cite{Stoica04}:
\begin{align}
\label{eq4}
&\notag \log p(\mathbf{Y}|\hat{\mathbf{B}}_t,\hat{\boldsymbol{\Sigma}}_t) = -\frac{M_2}{2}\log|\hat{\boldsymbol{\Sigma}}_t|\\ 
&-\frac{1}{2}\text{Tr}\{(\mathbf{Y}-\hat{\mathbf{P}}_t\hat{\mathbf{B}}_t)^T\hat{\boldsymbol{\Sigma}}_t^{-1}(\mathbf{Y}-\hat{\mathbf{P}}_t\hat{\mathbf{B}}_t)\} + \text{constant}.
\end{align} 
\subsection{ML estimation of the regression parameters} \label{ssec:MLest}
According to the PDM definition, the columns of $\mathbf{B}_t$, denoted by $\mathbf{b}_t^{(m)}$ for $m=1,\ldots,M_2$, are contained in a set $\mathbb{B}(\hat{\boldsymbol{\lambda}}_t )$. Thus, the ML estimate of $\mathbf{B}_t$ is also constrained. 
We define $\mathbb{B}(\hat{\boldsymbol{\lambda}}_t ) = \{\mathbf{b}_t\in \mathbb{R}^{t \times 1} : |b_i| \leq \sqrt{\hat{\lambda}_i}, \forall
 i = 1 \dots t\}$, where $\mathbf{b}_t = [b_1, \dots, b_t]^T$, and $\hat{\boldsymbol{\lambda}}_t = [\hat{\lambda}_1, \dots, \hat{\lambda}_t]^T$ contains the $t$ largest eigenvalues of $\hat{\mathbf{R}}_{yy}$. Then, the ML estimation of $\mathbf{B}_t$ is equivalent to the following regularized least-squares minimization \cite{Bedrick94}:
\begin{equation}
\begin{aligned}\label{eq2}
&\hat{\mathbf{B}}_t = \underset{{\mathbf{B}}_t \in \mathbb{B}(\hat{\boldsymbol{\lambda}}_t )}{\arg\min} 
& &\text{Tr}\{(\mathbf{Y}- \hat{\mathbf{P}}_t\mathbf{B}_t)^T\boldsymbol{\Sigma}_t^{-1}(\mathbf{Y}- \hat{\mathbf{P}}_t\mathbf{B}_t)\} ,\\
\end{aligned} 
\end{equation}
where $\mathbf{B}_t\in\mathbb{B}(\hat{\boldsymbol{\lambda}}_t)$ is applied column-wise. We calculate the solution to \eqref{eq2} following the lines of \cite{Cootes96}. That is, we first obtain the unconstrained solution to \eqref{eq2}, i.e., $ \hat{\mathbf{B}}^{u}_t = (\mathbf{P}_t^T\boldsymbol{\Sigma}_{t}^{-1}\mathbf{P}_t)^{-1}(\mathbf{P}^T_t\boldsymbol{\Sigma}_{_t}^{-1})\mathbf{Y}$. Then, we scale it such that the constraints are fullfilled.

The ML estimate of the covariance matrix of the residual noise $\boldsymbol{\Sigma}_t$ is, as long as $N<M_2$, the sample covariance matrix: 
\begin{equation}\label{MLnoise}
\hat{\boldsymbol{\Sigma}}_t =\frac{1}{M_2}(\mathbf{Y}- \hat{\mathbf{P}}_t\mathbf{B}_t)(\mathbf{Y}- \hat{\mathbf{P}}_t\mathbf{B}_t)^T.
\end{equation}
We observe that the ML estimates $\hat{\mathbf{B}}_t $ and  $\hat{\boldsymbol{\Sigma}}_t$ are mutually dependent, which prohibits finding a closed-form solution. We propose an alternating optimization algorithm to find a locally optimum solution. We set $\hat{\boldsymbol{\Sigma}}_t=\mathbf{I}$ as the initial point. Then, for each iteration of the alternating optimization, we solve \eqref{eq2} to calculate the ML estimate of $\hat{\mathbf{B}}_t$ and then, re-estimate the covariance matrix as in \eqref{MLnoise}. We repeat this procedure until convergence.

\subsection{Choosing the model order}
 The shape data is often high-dimensional but with small number of samples. Under these circumstances, the estimate of the matrix $\boldsymbol{\Sigma}_t$ may be ill-conditioned. In order to deal with this, we reduce the number of parameters to be estimated by assuming $\boldsymbol{\Sigma}_t$ to be a diagonal matrix, with $\boldsymbol{\sigma}^2 = [\sigma^2_1,\sigma^2_2,\dots,\sigma^2_N]$ on the diagonal. Consequently, its ML estimate is $\hat{\boldsymbol{\Sigma}}_t = \text{diag}(\frac{1}{M_2}(\mathbf{Y}- \hat{\mathbf{P}}_t\mathbf{B}_t)(\mathbf{Y}- \hat{\mathbf{P}}_t\mathbf{B}_t)^T)$. 
 Following the lines of \cite{Bedrick94}, we choose the Akaike information criterion (AIC), so the penalty term in \eqref{eq3} is $\eta(t) = M_2t + N$, which corresponds to the degrees of freedom in \eqref{eq4}. Notice that AIC has been used as well for similar problems (for example in \cite{Nadakuditi08}), and that the assumptions made to derive it are quite weak \cite{Stoica04}.
  Finally, the model-order estimate $t^*$ is obtained by minimizing the terms in \eqref{eq3} that depend on $t$: 
\begin{multline}
t^* = \arg\min_{t}\Big[  M_2\Big(\sum_{i=1}^{N}\log(\hat{\sigma}_i^2) +2t\Big) \\
+ \sum_{i=1}^{N}\sum_{m=1}^{M_2}\frac{\hat{\epsilon}^{2(m)}_i}{\hat{\sigma}_i^2}   \Big],
\end{multline}
where $\hat{\epsilon}^{(m)}_i$ is the $i$th element of vector $\hat{\boldsymbol{\epsilon}}^{(m)}= \mathbf{y}^{(m)}-\hat{\mathbf{P}}_t\hat{\mathbf{b}}^{(m)}_t$; and $\hat{\mathbf{b}}_t$ and  $\hat{\sigma}_i^2$ are the estimated values after convergence of the alternating optimization.

\label{ssec:WLS}
\section{Results and discussion}

In this section we validate our strategy with numerical results. In a PDM trained from real data sets, the true model order of the shape model is not known. Hence, we have also used simulated shape data following the model in Eq. \eqref{yP}, where the model order $t$ is set beforehand and thus known.
\subsection{Data description}\label{ssec:DataDes}

We simulate realistic synthetic shapes following the model in \eqref{yP}, similarly to the simulated data in \cite{Mei08}. We use the  eigenvectors $\hat{\mathbf{P}}_t$ obtained from available real shape data sets and choose values for $\mathbf{b}_t$ that are consistent with the sample data. Then we add white Gaussian noise $\boldsymbol{\epsilon}$ with different noise levels $\beta$, which we define as the ratio between the smallest kept signal eigenvalue and the noise variance. We choose these noise levels $\beta$ such that the produced shapes still look realistic. Then we randomly rotate, scale, and translate these synthetic shapes and use Procrustes to re-align them. Procrustes alignment typically colors the noise, so it may not longer be white when the model order has to be selected.

We have used the following databases of real 2D shapes:
\begin{itemize}
\item The top part of the femur bone as seen in fluoroscopic X-rays \cite{Eguizabal17} (168 samples with 40 landmarks).
\item Lung outlines from chest X-rays \cite{Shiraishi00} (246 samples with 44 landmarks). 
\item Hand outlines from photographs \cite{Stegmann02} (38 samples with 20 landmarks).
\end{itemize}

\begin{figure}[t] 
\center
		\includegraphics[scale=0.55]{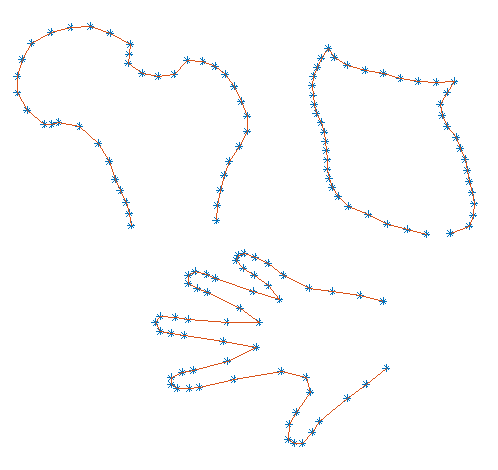}

		\caption{Average shapes of femur, lung, and hand. The landmarks are shown as blue stars.}
		\label{fig:Data_ex}
\end{figure}
We show the average shape of each data set in Fig. \ref{fig:Data_ex}. 

These three data sets are similar: they belong to human anatomy shapes, each sample contains a fixed number of landmarks in correspondence, and the noise is unknown. In each data set, the shapes are composed by a few landmarks that are anatomical  and manually labeled, and the rest of the landmarks are equally distributed between these. We refer to the papers \cite{Eguizabal17},\cite{Shiraishi00} and \cite{Stegmann02} for further details about these data sets.

\subsection{Evaluation of the results}
We compare our proposed strategy with four others: 
\begin{enumerate}
\item A variance threshold of 95\%, as described in \cite{Cootes96} and validated in \cite{Lindner13}.
\item An information-theoretic technique that considers white noise \cite{Nadakuditi08}.
\item An information-theoretic criterion considering non-uniform noise \cite{Aouada04}.
\item A boostrap t-test designed for PDMs \cite{Mei08} that considers white noise.
\end{enumerate}
\subsection{Simulated data}
The evaluation of simulated data is straightforward since there is a known ground truth for the model order $t$.

\begin{figure}[t] 
\center
\begin{subfigure}[b]{0.5\textwidth}
\center
%
%
\definecolor{mycolor1}{rgb}{0.00000,0.44700,0.74100}%
\definecolor{mycolor2}{rgb}{0.85000,0.32500,0.09800}%
\definecolor{mycolor3}{rgb}{0.92900,0.69400,0.12500}%
\definecolor{mycolor4}{rgb}{0.49400,0.18400,0.55600}%
\definecolor{mycolor5}{rgb}{0.46600,0.67400,0.18800}%
\definecolor{mycolor6}{rgb}{0.30100,0.74500,0.93300}%
\begin{tikzpicture}
\begin{axis}[%
width= \figurewidth,
height = \figureheight,
at={(0in,0in)},
scale only axis,
xmode=log,
xmin=10,
xmax=500,
xminorticks=true,
xtick={10,100,500}, xticklabels={10, 100, 500},
xlabel style={yshift=0.25cm,font=\color{white!9!black}},
xlabel={number of samples ($M$)},
ymin=0,
ymax=20,
ylabel style={yshift=-0.5cm,font=\color{white!9!black}},
ylabel={selected $t^*$},
axis background/.style={fill=white},
legend style={legend cell align=left, at={(0.5,1.03)},legend columns=2, column sep=1, fill=none, align=left, anchor=south, draw=none, font=9}
]

\addplot [color=mycolor1,legend image post style={mark=square*}]
  table[row sep=crcr]{%
10	4\\
15	6\\
20	7\\
25	8\\
30	8\\
35	8\\
40	9\\
45	9\\
50	9\\
55	9\\
60	9\\
65	9\\
70	9\\
75	9\\
80	9\\
85	9\\
90	10\\
95	10\\
100	10\\
200	10\\
300	10\\
400	10\\
500	10\\
};
\addlegendentry{proposed}
\addplot [color=mycolor2, draw=none, mark=*, mark options={solid, mycolor2}, forget plot]
 plot [error bars/.cd, y dir = both, y explicit]
 table[row sep=crcr, y error plus index=2, y error minus index=3]{%
10	1	0.334459322711673	0.334459322711673\\
20	5	0.569127310618291	0.569127310618291\\
30	8	0.494274703978666	0.494274703978666\\
40	10	0.463044494710425	0.463044494710425\\
50	11	0.647306647340817	0.647306647340817\\
60	12	0.823811725843465	0.823811725843465\\
70	12	0.780669083342664	0.780669083342664\\
80	11	0.683980778396162	0.683980778396162\\
90	11	0.614411393566408	0.614411393566408\\
100	11	0.580188608007497	0.580188608007497\\
300	14	0.665006265634637	0.665006265634637\\
500	23	1.04892386274756	1.04892386274756\\
};
\addplot [color=mycolor3,legend image post style={mark=triangle*}]
  table[row sep=crcr]{%
10	8\\
15	11\\
20	14\\
25	17\\
30	20\\
35	22\\
40	24\\
45	26\\
50	28\\
55	29\\
60	31\\
65	32\\
70	33\\
75	35\\
80	36\\
85	37\\
90	38\\
95	39\\
100	39\\
200	48\\
300	52\\
400	54\\
500	56\\
};
\addlegendentry{95\% \cite{Lindner13}}

\addplot [color=mycolor1, draw=none, mark=square*, mark options={solid, mycolor1}, forget plot]
 plot [error bars/.cd, y dir = both, y explicit]
 table[row sep=crcr, y error plus index=2, y error minus index=3]{%
10	4	0.362022537227058	0.362022537227058\\
20	7	0.735479235067458	0.735479235067458\\
30	8	0.814887691717931	0.814887691717931\\
40	9	0.716023916802448	0.716023916802448\\
50	9	0.574689307759866	0.574689307759866\\
60	9	0.468297684265501	0.468297684265501\\
70	9	0.409760728810468	0.409760728810468\\
80	9	0.385307251429991	0.385307251429991\\
90	10	0.341578634751679	0.341578634751679\\
100	10	0.304199257409709	0.304199257409709\\
300	10	0.0509078577169011	0.0509078577169011\\
500	10	0	0\\
};
\addplot [color=mycolor2,legend image post style={mark=*}]
  table[row sep=crcr]{%
10	1\\
15	3\\
20	5\\
25	7\\
30	8\\
35	9\\
40	10\\
45	10\\
50	11\\
55	11\\
60	12\\
65	12\\
70	12\\
75	11\\
80	11\\
85	11\\
90	11\\
95	11\\
100	11\\
200	12\\
300	14\\
400	18\\
500	23\\
};
\addlegendentry{white noise \cite{Nadakuditi08}}

\addplot [color=mycolor3, draw=none, mark=triangle*, mark options={solid, mycolor3}, forget plot]
 plot [error bars/.cd, y dir = both, y explicit]
 table[row sep=crcr, y error plus index=2, y error minus index=3]{%
10	8	0.176886655485621	0.176886655485621\\
20	14	0.253369134598804	0.253369134598804\\
30	20	0.316376881370556	0.316376881370556\\
40	24	0.355971412136572	0.355971412136572\\
50	28	0.358885905226886	0.358885905226886\\
60	31	0.394233088275396	0.394233088275396\\
70	33	0.398806609797456	0.398806609797456\\
80	36	0.405098179639692	0.405098179639692\\
90	38	0.407299886175913	0.407299886175913\\
100	39	0.399672201285909	0.399672201285909\\
300	52	0.283044670881267	0.283044670881267\\
500	56	0.214784604438623	0.214784604438623\\
};
\addplot [color=mycolor4,legend image post style={mark=diamond*}]
  table[row sep=crcr]{%
10	9\\
15	14\\
20	19\\
25	1\\
30	1\\
35	1\\
40	1\\
45	1\\
50	2\\
55	2\\
60	2\\
65	2\\
70	2\\
75	2\\
80	2\\
85	2\\
90	2\\
95	2\\
100	2\\
200	4\\
300	4\\
400	4\\
500	5\\
};
\addlegendentry{nonuniform \cite{Aouada04}}

\addplot [color=mycolor4, draw=none, mark=diamond*, mark options={solid, mycolor4}, forget plot]
 plot [error bars/.cd, y dir = both, y explicit]
 table[row sep=crcr, y error plus index=2, y error minus index=3]{%
10	9	0	0\\
20	19	0	0\\
30	1	0.197085337824729	0.197085337824729\\
40	1	0.244450886330287	0.244450886330287\\
50	2	0.243893679672367	0.243893679672367\\
60	2	0.230177393879287	0.230177393879287\\
70	2	0.20745952805864	0.20745952805864\\
80	2	0.228074857425298	0.228074857425298\\
90	2	0.249000268648144	0.249000268648144\\
100	2	0.267863386170627	0.267863386170627\\
300	4	0.0650204921474216	0.0650204921474216\\
500	5	0.409129378158745	0.409129378158745\\
};
\addplot [color=mycolor5,legend image post style={mark=star}]
  table[row sep=crcr]{%
10	2\\
20	2\\
30	3\\
40	1\\
50	2.5\\
60	3\\
70	3\\
80	3\\
90	3\\
100	3\\
200	4\\
300	4\\
400	5\\
500	5\\
};
\addlegendentry{t-test \cite{Mei08}}

\addplot [color=mycolor5, draw=none, mark=star, mark options={solid, mycolor5}, forget plot]
 plot [error bars/.cd, y dir = both, y explicit]
 table[row sep=crcr, y error plus index=2, y error minus index=3]{%
10	2	0.377305157795076	0.377305157795076\\
30	3	0.481958453206587	0.481958453206587\\
50	2.5	7.15162490709129	7.15162490709129\\
70	3	4.94251832975727	4.94251832975727\\
90	3	3.3515703534342	3.3515703534342\\
200	4	12.0365248822668	12.0365248822668\\
400	5	9.4301291177063	9.4301291177063\\
};
\addplot [color=gray, dashed]
  table[row sep=crcr]{%
10	10\\
15	10\\
20	10\\
25	10\\
30	10\\
35	10\\
40	10\\
45	10\\
50	10\\
55	10\\
60	10\\
65	10\\
70	10\\
75	10\\
80	10\\
85	10\\
90	10\\
95	10\\
100	10\\
200	10\\
300	10\\
400	10\\
500	10\\
};
\addlegendentry{true order $t$ }
\end{axis}
\end{tikzpicture}%
	\caption{}
\end{subfigure}%
\label{fig:5dB}
\begin{subfigure}[b]{0.5\textwidth}
\center
%
%
\definecolor{mycolor1}{rgb}{0.00000,0.44700,0.74100}%
\definecolor{mycolor2}{rgb}{0.85000,0.32500,0.09800}%
\definecolor{mycolor3}{rgb}{0.92900,0.69400,0.12500}%
\definecolor{mycolor4}{rgb}{0.49400,0.18400,0.55600}%
\definecolor{mycolor5}{rgb}{0.46600,0.67400,0.18800}%
\definecolor{mycolor6}{rgb}{0.30100,0.74500,0.93300}%
\begin{tikzpicture}

\begin{axis}[%
width= \figurewidth,
height = \figureheight,
at={(0in,0in)},
scale only axis,
xmode=log,
xmin=10,
xmax=500,
xminorticks=true,
xtick={10,100,500}, xticklabels={10, 100, 500},
xlabel style={yshift=0.25cm,font=\color{white!9!black}},
xlabel={number of samples ($M$)},
ymin=0,
ymax=20,
ylabel style={yshift=-0.5cm,font=\color{white!9!black}},
ylabel={selected $t^*$},
axis background/.style={fill=white},
]
\addplot [color=mycolor1]
  table[row sep=crcr]{%
10	4\\
15	6\\
20	8\\
25	10\\
30	10\\
35	10\\
40	10\\
45	10\\
50	10\\
55	10\\
60	10\\
65	10\\
70	10\\
75	10\\
80	10\\
85	10\\
90	10\\
95	10\\
100	10\\
200	10\\
300	10\\
400	10\\
500	10\\
};

\addplot [color=mycolor1, draw=none, mark=square*, mark options={solid, mycolor1}, forget plot]
 plot [error bars/.cd, y dir = both, y explicit]
 table[row sep=crcr, y error plus index=2, y error minus index=3]{%
10	4	0.0923674511489082	0.0923674511489082\\
20	8	0.293319804882841	0.293319804882841\\
30	10	0.200690983226203	0.200690983226203\\
40	10	0.0613602425441692	0.0613602425441692\\
50	10	0.0154229854673347	0.0154229854673347\\
60	10	0	0\\
70	10	0	0\\
80	10	0	0\\
90	10	0	0\\
100	10	0	0\\
300	10	0	0\\
500	10	0	0\\
};
\addplot [color=mycolor2]
  table[row sep=crcr]{%
10	1\\
15	2\\
20	10\\
25	10\\
30	10\\
35	10\\
40	10\\
45	11\\
50	11\\
55	12\\
60	12\\
65	12\\
70	12\\
75	12\\
80	12\\
85	11\\
90	11\\
95	11\\
100	11\\
200	12\\
300	14\\
400	18\\
500	23\\
};

\addplot [color=mycolor2, draw=none, mark=*, mark options={solid, mycolor2}, forget plot]
 plot [error bars/.cd, y dir = both, y explicit]
 table[row sep=crcr, y error plus index=2, y error minus index=3]{%
10	1	0.45333758501408	0.45333758501408\\
20	10	0.0154229854673348	0.0154229854673348\\
30	10	0.137664352404325	0.137664352404325\\
40	10	0.396834412595995	0.396834412595995\\
50	11	0.656301220871185	0.656301220871185\\
60	12	0.898074155183531	0.898074155183531\\
70	12	0.827345014635747	0.827345014635747\\
80	12	0.702208372664026	0.702208372664026\\
90	11	0.607176282926753	0.607176282926753\\
100	11	0.587552091577808	0.587552091577808\\
300	14	0.693092579864363	0.693092579864363\\
500	23	1.01481363413099	1.01481363413099\\
};
\addplot [color=mycolor3]
  table[row sep=crcr]{%
10	6\\
15	7\\
20	7\\
25	8\\
30	8\\
35	8\\
40	8\\
45	8\\
50	8\\
55	8\\
60	8\\
65	8\\
70	8\\
75	8\\
80	8\\
85	8\\
90	8\\
95	9\\
100	9\\
200	9\\
300	9\\
400	9\\
500	9\\
};

\addplot [color=mycolor3, draw=none, mark=triangle*, mark options={solid, mycolor3}, forget plot]
 plot [error bars/.cd, y dir = both, y explicit]
 table[row sep=crcr, y error plus index=2, y error minus index=3]{%
10	6	0.259161389582515	0.259161389582515\\
20	7	0.231806960896736	0.231806960896736\\
30	8	0.197318187504806	0.197318187504806\\
40	8	0.140172391431864	0.140172391431864\\
50	8	0.181412947708732	0.181412947708732\\
60	8	0.219317121994612	0.219317121994612\\
70	8	0.237535055295699	0.237535055295699\\
80	8	0.245422953905813	0.245422953905813\\
90	8	0.249799919935936	0.249799919935936\\
100	9	0.248193472919818	0.248193472919818\\
300	9	0.0872364836615552	0.0872364836615552\\
500	9	0.0218009964614469	0.0218009964614469\\
};
\addplot [color=mycolor4]
  table[row sep=crcr]{%
10	9\\
15	14\\
20	19\\
25	2\\
30	3\\
35	3\\
40	3\\
45	4\\
50	4\\
55	4\\
60	4\\
65	5\\
70	8\\
75	10\\
80	10\\
85	10\\
90	10\\
95	10\\
100	10\\
200	10\\
300	10\\
400	10\\
500	10\\
};

\addplot [color=mycolor4, draw=none, mark=diamond*, mark options={solid, mycolor4}, forget plot]
 plot [error bars/.cd, y dir = both, y explicit]
 table[row sep=crcr, y error plus index=2, y error minus index=3]{%
10	9	0	0\\
20	19	0	0\\
30	3	0.353384991731329	0.353384991731329\\
40	3	0.347130772537714	0.347130772537714\\
50	4	0.341773751262664	0.341773751262664\\
60	4	0.492953521747413	0.492953521747413\\
70	8	1.16267538847131	1.16267538847131\\
80	10	0.264193033942369	0.264193033942369\\
90	10	0.0701294559337887	0.0701294559337887\\
100	10	0.0376883027379229	0.0376883027379229\\
300	10	0	0\\
500	10	0	0\\
};
\addplot [color=mycolor5]
  table[row sep=crcr]{%
10	2\\
20	3\\
30	4\\
40	1\\
50	1\\
60	1\\
70	1\\
80	1\\
90	1\\
100	1\\
200	12\\
300	11\\
400	13\\
500	12\\
};

\addplot [color=mycolor5, draw=none, mark=star, mark options={solid, mycolor5}, forget plot]
 plot [error bars/.cd, y dir = both, y explicit]
 table[row sep=crcr, y error plus index=2, y error minus index=3]{%
10	2	0.415943559782519	0.415943559782519\\
30	4	0.766797353103818	0.766797353103818\\
50	1	8.61823031826067	8.61823031826067\\
70	1	7.4397867402499	7.4397867402499\\
90	1	6.6901617236722	6.6901617236722\\
200	12	15.8667926125716	15.8667926125716\\
400	13	16.3426267901751	16.3426267901751\\
};
\addplot [color=gray, dashed]
  table[row sep=crcr]{%
10	10\\
15	10\\
20	10\\
25	10\\
30	10\\
35	10\\
40	10\\
45	10\\
50	10\\
55	10\\
60	10\\
65	10\\
70	10\\
75	10\\
80	10\\
85	10\\
90	10\\
95	10\\
100	10\\
200	10\\
300	10\\
400	10\\
500	10\\
};

\end{axis}
\end{tikzpicture}%
		\caption{}
\label{fig:20dB}
\end{subfigure}%
		\caption{Average model order over 1000 Monte Carlo simulations, using simulated femur shapes of 40 landmarks, with different number of samples $M$. The vertical bars indicate variance. True model order is $t=10$. We consider two different signal-to-noise levels: (a) $\beta=5$ dB. (b) $\beta=20$ dB. 
}
		\label{fig:Art}
\end{figure}
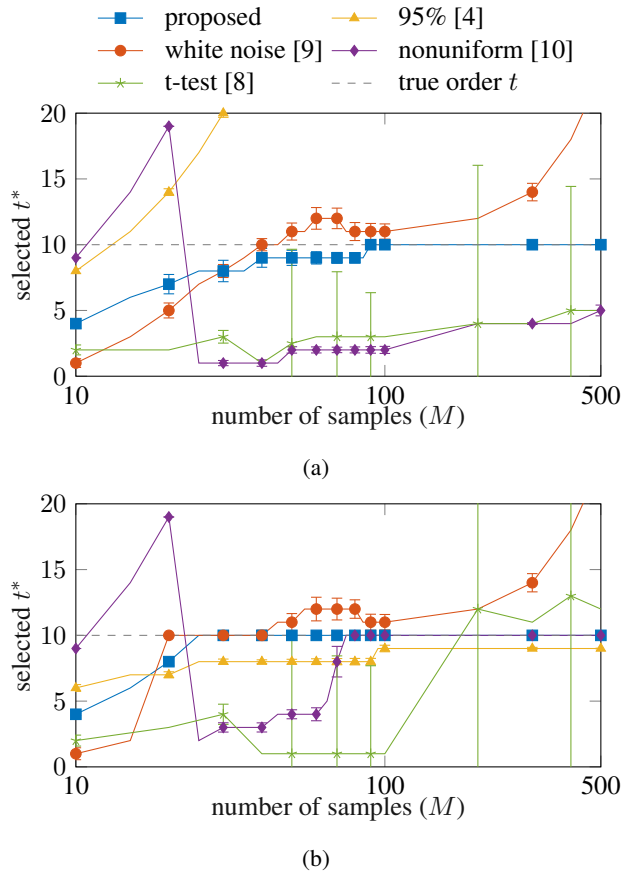

In Fig. \ref{fig:Art} we show the model order obtained from 1000 Monte Carlo simulations. We see that the performance of the 95\%-approach  \cite{Lindner13} (triangle, yellow lines) depends considerably on the level of noise: there is a tendency to overestimate if $\beta$ is moderate (5 dB) and to underestimate if $\beta$ is high (20 dB). The white-noise strategy in \cite{Nadakuditi08} (circles, red lines) tends to overestimate, especially when the number of samples increases. We believe this is due to color in the noise, introduced in the simulation by the Procrustes alignment. The t-test stategy \cite{Mei08} (stars, green lines) does not perform well and leads to results with high variance. The reason for this may be that this approach assumes white noise to evaluate stability. The nonuniform noise strategy \cite{Aouada04} (diamond, purple lines) tends to underestimate if $\beta$ is moderate (5 dB), and it provides an incorrect estimation if the number of samples is very small. 
Our proposed strategy (square, blue lines) outperforms the competitors: it provides the best model-order estimate in general, it needs fewer samples to find the correct estimate, it is not highly dependent on the noise level, and it is consistent with increasing number of samples.

\subsection{Real data}
There is no known ground truth for model order $t$ in a PDM that is trained with real shape data sets. Nevertheless, we may still evaluate how plausible the model-order estimate is when the number of available samples changes and compare this with the behavior in artificial data. Additionally, we illustrate the importance of the model order in PDMs with a numerical experiment that shows the impact of the selection of $t$ when a PDM of order $t$ is used to deal with partial occlusions in shapes.

\begin{figure}[t] 
\center
\begin{subfigure}[b]{0.5\textwidth}
\center
%
%
\definecolor{mycolor1}{rgb}{0.00000,0.44700,0.74100}%
\definecolor{mycolor2}{rgb}{0.85000,0.32500,0.09800}%
\definecolor{mycolor3}{rgb}{0.92900,0.69400,0.12500}%
\definecolor{mycolor4}{rgb}{0.49400,0.18400,0.55600}%
\definecolor{mycolor5}{rgb}{0.46600,0.67400,0.18800}%
\begin{tikzpicture}

\begin{axis}[%
width= \figurewidth,
height = \figureheight/1.5,
at={(0in,0in)},
scale only axis,
xmin=10,
xmax=165,
xlabel style={yshift=0.25cm,font=\color{white!9!black}},
xlabel={number of samples ($M$)},
ymin=0,
ymax=20,
ylabel style={yshift=-0.5cm,font=\color{white!9!black}},
ylabel={selected $t^*$},
axis background/.style={fill=white},
title style={font=\bfseries},
legend style={legend cell align=left, at={(0.5,1.03)},legend columns=2, column sep=1, fill=none, align=left, anchor=south, draw=none, font=9}
]

\addplot [color=mycolor1, draw=none, mark=square*]
  table[row sep=crcr]{%
10	4\\
20	9\\
40	13\\
60	14\\
80	15\\
100	15\\
120	15\\
140	15\\
160	15\\
};
\addlegendentry{proposed}
\addplot [color=mycolor2,draw=none, mark=* ]
  table[row sep=crcr]{%
10	2\\
20	7\\
40	20\\
60	30.5\\
80	52\\
120	49\\
140	51\\
160	52\\
};

\addlegendentry{white noise \cite{Nadakuditi08}}

\addplot [color=mycolor3, draw=none, mark=triangle*]
  table[row sep=crcr]{%
10	6\\
20	9\\
40	11\\
60	12\\
80	12\\
100	13\\
120	13\\
140	13\\
160	13\\
};
\addlegendentry{95\% \cite{Lindner13}}

\addplot [color=mycolor4, draw=none,mark=diamond*]
  table[row sep=crcr]{%
10	9\\
20	19\\
40	2\\
60	3\\
80	4\\
100	4\\
120	4\\
140	4\\
160	4\\
};
\addlegendentry{nonuniform \cite{Aouada04}}
\addplot [color=mycolor5, dashed, forget plot]
  table[row sep=crcr]{%
10	2\\
15	2\\
20	2\\
25	2\\
30	3\\
35	4.5\\
40	5\\
45	5\\
50	5\\
55	4.5\\
60	4.5\\
65	5\\
70	4\\
75	5\\
80	5\\
85	5.5\\
90	5\\
95	5\\
100	4\\
105	6\\
110	6\\
115	6\\
120	6\\
125	7\\
130	7\\
135	8\\
140	7\\
145	8\\
150	9\\
155	8\\
160	9\\
165	10\\
};

\addplot [color=mycolor5, draw=none,mark=star]
  table[row sep=crcr]{%
10	2\\
20	2\\
40	5\\
60	4.5\\
80	5\\
100	4\\
120	6\\
140	7\\
160	9\\
};
\addlegendentry{t-test \cite{Mei08}}
\addplot [color=mycolor1, dashed]
  table[row sep=crcr]{%
10	4\\
15	6\\
20	9\\
25	10\\
30	11\\
35	12\\
40	13\\
45	13\\
50	14\\
55	14\\
60	14\\
65	14\\
70	14\\
75	14\\
80	15\\
85	15\\
90	14\\
95	15\\
100	15\\
105	15\\
110	15\\
115	15\\
120	15\\
125	15\\
130	15\\
135	15\\
140	15\\
145	15\\
150	15\\
155	15\\
160	15\\
165	15\\
};
\addplot [color=mycolor2, dashed]
  table[row sep=crcr]{%
10	2\\
15	3\\
20	7\\
25	13\\
30	15\\
35	18\\
40	20\\
45	22\\
50	25\\
55	28\\
60	30.5\\
65	35\\
70	40.5\\
75	46\\
80	52\\
85	51\\
90	49\\
95	49\\
100	48\\
105	49\\
110	49\\
115	49\\
120	49\\
125	49\\
130	50\\
135	51\\
140	51\\
145	51\\
150	52\\
155	52\\
160	52\\
165	52\\
};
\addplot [color=mycolor3, dashed]
  table[row sep=crcr]{%
10	6\\
15	8\\
20	9\\
25	10\\
30	10\\
35	11\\
40	11\\
45	11\\
50	12\\
55	12\\
60	12\\
65	12\\
70	12\\
75	12\\
80	12\\
85	12\\
90	12\\
95	13\\
100	13\\
105	13\\
110	13\\
115	13\\
120	13\\
125	13\\
130	13\\
135	13\\
140	13\\
145	13\\
150	13\\
155	13\\
160	13\\
165	13\\
};
\addplot [color=mycolor4, dashed]
  table[row sep=crcr]{%
10	9\\
15	14\\
20	19\\
25	2\\
30	2\\
35	2\\
40	2\\
45	3\\
50	3\\
55	3\\
60	3\\
65	4\\
70	4\\
75	4\\
80	4\\
85	4\\
90	4\\
95	4\\
100	4\\
105	4\\
110	4\\
115	4\\
120	4\\
125	4\\
130	4\\
135	4\\
140	4\\
145	4\\
150	4\\
155	4\\
160	4\\
165	4\\
};

\end{axis}
\end{tikzpicture}%
		 	\caption{}
\end{subfigure}
\begin{subfigure}[b]{0.5\textwidth}
\center
%
%
\definecolor{mycolor1}{rgb}{0.00000,0.44700,0.74100}%
\definecolor{mycolor2}{rgb}{0.85000,0.32500,0.09800}%
\definecolor{mycolor3}{rgb}{0.92900,0.69400,0.12500}%
\definecolor{mycolor4}{rgb}{0.49400,0.18400,0.55600}%
\definecolor{mycolor5}{rgb}{0.46600,0.67400,0.18800}%
\begin{tikzpicture}

\begin{axis}[%
width= \figurewidth,
height = \figureheight/1.5,
at={(0in,0in)},
scale only axis,
unbounded coords=jump,
xmin=10,
xmax=245,
xlabel style={yshift=0.25cm,font=\color{white!9!black}},
xlabel={number of samples ($M$)},
ymin=0,
ymax=20,
ylabel style={yshift=-0.5cm,font=\color{white!9!black}},
ylabel={selected $t^*$},
axis background/.style={fill=white},
title style={font=\bfseries},
]
\addplot [color=mycolor1, dashed]
  table[row sep=crcr]{%
10	4\\
15	6\\
20	8\\
25	9\\
30	9.5\\
35	9.5\\
40	10\\
45	10\\
50	10.5\\
55	10.5\\
60	11\\
65	10\\
70	10\\
75	11\\
80	10\\
85	10\\
90	10.5\\
95	10\\
100	10\\
105	10\\
110	10\\
115	10\\
120	10\\
125	10\\
130	10\\
135	10\\
140	10\\
145	10\\
150	10\\
155	10\\
160	10\\
165	10\\
170	10\\
175	11\\
180	11\\
185	10\\
190	10\\
195	10\\
200	10\\
205	10\\
210	10\\
215	10\\
220	11\\
225	10\\
230	10\\
235	10\\
240	10\\
245	nan\\
};

\addplot [color=mycolor1, draw=none, mark=square*]
  table[row sep=crcr]{%
10	4\\
50	10\\
100	10\\
150	10\\
200	10\\
240	10\\
};%

\addplot [color=mycolor2, dashed]
  table[row sep=crcr]{%
10	3\\
15	5\\
20	8\\
25	11\\
30	15\\
35	17\\
40	20\\
45	22\\
50	24\\
55	26\\
60	29\\
65	32\\
70	33\\
75	34\\
80	69\\
85	74\\
90	78\\
95	77\\
100	76\\
105	77\\
110	76.5\\
115	77\\
120	76.5\\
125	76\\
130	77\\
135	76\\
140	76\\
145	77\\
150	76\\
155	75\\
160	75.5\\
165	84\\
170	84\\
175	84.5\\
180	85\\
185	85\\
190	85\\
195	85\\
200	86\\
205	86\\
210	86\\
215	86\\
220	86\\
225	86\\
230	86\\
235	86\\
240	86\\
245	nan\\
};

\addplot [color=mycolor2, draw=none, mark=*]
  table[row sep=crcr]{%
10	3\\
35	17\\
100	76\\
150	76\\
200	86\\
240	86\\
};

\addplot [color=mycolor3, dashed]
  table[row sep=crcr]{%
10	5\\
15	6\\
20	6.5\\
25	7\\
30	7\\
35	7\\
40	8\\
45	8\\
50	8\\
55	8\\
60	8\\
65	8\\
70	8\\
75	8\\
80	8\\
85	8\\
90	8\\
95	8\\
100	8\\
105	8\\
110	8\\
115	8\\
120	8\\
125	8\\
130	8\\
135	8\\
140	8\\
145	8\\
150	8\\
155	8\\
160	8\\
165	8\\
170	8\\
175	8\\
180	8\\
185	8\\
190	8\\
195	8\\
200	8\\
205	8\\
210	8\\
215	8\\
220	8\\
225	8\\
230	8\\
235	8\\
240	8\\
245	nan\\
};

\addplot [color=mycolor3,draw=none, mark=triangle*]
  table[row sep=crcr]{%
10	5\\
50	8\\
100	8\\
150	8\\
200	8\\
240	8\\
};

\addplot [color=mycolor4, dashed]
  table[row sep=crcr]{%
10	9\\
15	14\\
20	19\\
25	2\\
30	3\\
35	3\\
40	3\\
45	3\\
50	3\\
55	4\\
60	4\\
65	5\\
70	5\\
75	6\\
80	6\\
85	6\\
90	6\\
95	6\\
100	7\\
105	7\\
110	7\\
115	7\\
120	7\\
125	8\\
130	8\\
135	8\\
140	8\\
145	8\\
150	8\\
155	8\\
160	8\\
165	8\\
170	8\\
175	8\\
180	8\\
185	8\\
190	8\\
195	8\\
200	8\\
205	8\\
210	8\\
215	8\\
220	8\\
225	8\\
230	8\\
235	8\\
240	8\\
245	nan\\
};

\addplot [color=mycolor4, draw=none, mark=diamond*]
  table[row sep=crcr]{%
10	9\\
50	3\\
100	7\\
150	8\\
200	8\\
240	8\\
};

\addplot [color=mycolor5, dashed]
  table[row sep=crcr]{%
10	1\\
15	1\\
20	1\\
25	1\\
30	1\\
35	1\\
40	2\\
45	2\\
50	1\\
55	1\\
60	1\\
65	1\\
70	1\\
75	1\\
80	2\\
85	2\\
90	1\\
95	1\\
100	2\\
105	1.5\\
110	3\\
115	2\\
120	3\\
125	3\\
130	4\\
135	3\\
140	3\\
145	3\\
150	3\\
155	3\\
160	3\\
165	3\\
170	4\\
175	3\\
180	3\\
185	3\\
190	3\\
195	3\\
200	3\\
205	2\\
210	3\\
215	3\\
220	3\\
225	3\\
230	3\\
235	2\\
240	2\\
245	nan\\
};

\addplot [color=mycolor5,draw=none, mark=star ]
  table[row sep=crcr]{%
10	1\\
50	1\\
100	2\\
150	3\\
200	3\\
240	2\\
};

\end{axis}
\end{tikzpicture}%
			\caption{}
		\end{subfigure}
		\begin{subfigure}[b]{0.5\textwidth}
	\center
%
%
\definecolor{mycolor1}{rgb}{0.00000,0.44700,0.74100}%
\definecolor{mycolor2}{rgb}{0.85000,0.32500,0.09800}%
\definecolor{mycolor3}{rgb}{0.92900,0.69400,0.12500}%
\definecolor{mycolor4}{rgb}{0.49400,0.18400,0.55600}%
\definecolor{mycolor5}{rgb}{0.46600,0.67400,0.18800}%
\begin{tikzpicture}

\begin{axis}[%
width= \figurewidth,
height = \figureheight/1.5,
at={(0in,0in)},
scale only axis,
unbounded coords=jump,
xmin=10,
xmax=35,
xlabel style={yshift=0.25cm,font=\color{white!9!black}},
xlabel={number of samples ($M$)},
ymin=0,
ymax=20,
ylabel style={yshift=-0.5cm,font=\color{white!9!black}},
ylabel={selected $t^*$},
axis background/.style={fill=white},
title style={font=\bfseries},
]
\addplot [color=mycolor1,mark=square*, dashed]
  table[row sep=crcr]{%
10	4\\
15	6\\
20	8\\
25	9\\
30	9\\
35	9\\
40	nan\\
};

\addplot [color=mycolor2, mark=*, dashed]
  table[row sep=crcr]{%
10	3\\
15	6\\
20	9\\
25	11\\
30	13\\
35	15\\
40	nan\\
};

\addplot [color=mycolor3, mark=triangle*, dashed]
  table[row sep=crcr]{%
10	3\\
15	4\\
20	4\\
25	5\\
30	5\\
35	5\\
40	nan\\
};

\addplot [color=mycolor4, mark=diamond*, dashed]
  table[row sep=crcr]{%
10	9\\
15	14\\
20	19\\
25	3\\
30	3\\
35	3\\
40	nan\\
};

\addplot [color=mycolor5,mark=star, dashed]
  table[row sep=crcr]{%
10	1\\
15	1\\
20	1\\
25	1\\
30	2\\
35	2\\
40	nan\\
};

\end{axis}
\end{tikzpicture}%
			\caption{}
		\end{subfigure}
		\caption{Selected model order for three different real shape data sets: (a) femur,  (b) lung, and (c) hand. Notice that the different shapes have different number of total samples $M$. 
}
		\label{fig:Real}
\end{figure}
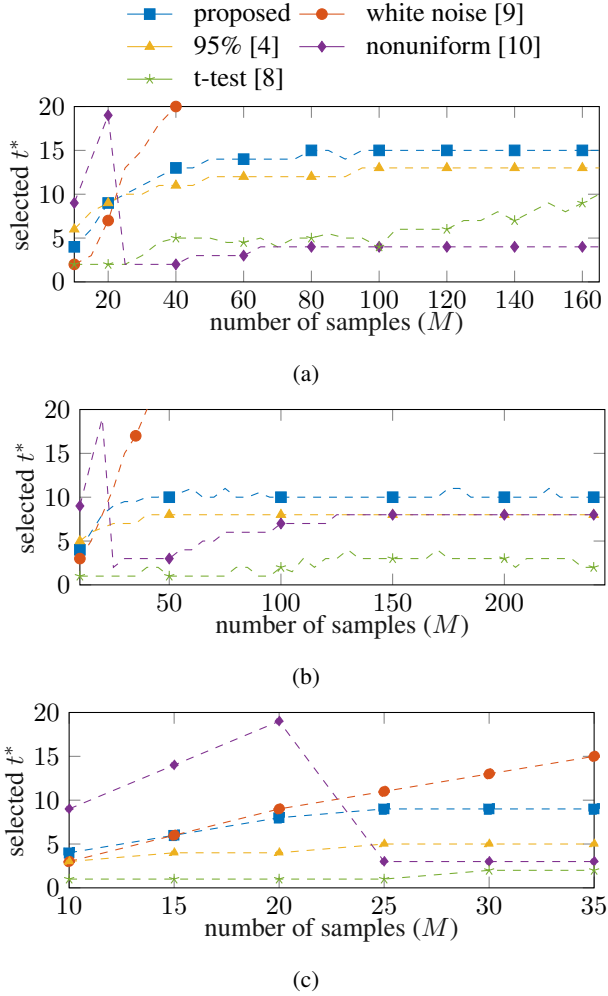 
 In Fig. \ref{fig:Real} we show the estimated model order for different number of samples on the three data sets. There are similarities with Fig. \ref{fig:Art}:  ``95\%" \cite{Lindner13}, t-test \cite{Mei08}, and ``nonuniform" \cite{Aouada04} provide small model orders, while ``white-noise" \cite{Nadakuditi08} provides large (probably too large) model orders. Our strategy seems to provide a consistent model order that starts converging with fewer samples. 

\begin{figure}[t] 
\center
%
%
\definecolor{mycolor1}{rgb}{0.30100,0.74500,0.93300}%
\definecolor{mycolor2}{rgb}{0.63500,0.07800,0.18400}%
\definecolor{mycolor3}{rgb}{0.00000,0.44700,0.74100}%
\definecolor{mycolor4}{rgb}{0.85000,0.32500,0.09800}%
\definecolor{mycolor5}{rgb}{0.92900,0.69400,0.12500}%
\definecolor{mycolor6}{rgb}{0.49400,0.18400,0.55600}%
\definecolor{Mei}{rgb}{0.46600,0.67400,0.18800}%
\begin{tikzpicture}

\begin{axis}[%
width= \figurewidth,
height = \figureheight,
at={(0in,0in)},
scale only axis,
unbounded coords=jump,
xmin=1,
xmax= 55,
xlabel style={yshift=0.25cm,font=\color{white!9!black}},
xlabel={Model order $t$},
ymin=0,
ymax=0.00003,
ylabel style={yshift=-0.5cm,font=\color{white!9!black}},
ylabel={ $e_{\text{LMMSE}}$ },
axis background/.style={fill=white},
legend style={legend cell align=left, at={(0.5,1.1)},legend columns=3, column sep=1, fill=none, align=left, anchor=south, draw=none, font=9}
]
\addplot [color=magenta, dotted]
  table[row sep=crcr]{%
1	3.31429273639752e-05\\
2	2.67309190257437e-05\\
3	2.23863029344807e-05\\
4	1.73371668410293e-05\\
5	1.53691460926249e-05\\
6	1.31816335108502e-05\\
7	1.12639252120729e-05\\
8	9.12360264188522e-06\\
9	7.95631136597516e-06\\
10	7.52204514376317e-06\\
11	5.96960609788651e-06\\
12	5.20120006482429e-06\\
13	4.69080179778324e-06\\
14	4.12319995485332e-06\\
15	3.75671986166275e-06\\
16	3.5720802525742e-06\\
17	3.52986217551485e-06\\
18	3.51838572564274e-06\\
19	3.35537715434709e-06\\
20	3.39479596415976e-06\\
21	3.31528837834596e-06\\
22	3.3278806625232e-06\\
23	3.36112956342217e-06\\
24	3.38209617247062e-06\\
25	3.40897178886117e-06\\
26	3.39590521788802e-06\\
27	3.45631919655496e-06\\
28	3.51872847907181e-06\\
29	3.60513470384402e-06\\
30	3.64422704398258e-06\\
31	3.72050816723681e-06\\
32	3.86987598548659e-06\\
33	4.01810816637904e-06\\
34	4.29513015437234e-06\\
35	4.5888405443502e-06\\
36	4.85653247820819e-06\\
37	5.10315102188297e-06\\
38	5.3990969157152e-06\\
39	5.66693341995009e-06\\
40	5.76818631113656e-06\\
41	6.23874203503313e-06\\
42	6.63423407737754e-06\\
43	6.94073723774323e-06\\
44	7.33386061737986e-06\\
45	7.85696104289247e-06\\
46	8.40908453415151e-06\\
47	9.50863360932878e-06\\
48	1.04033905618563e-05\\
49	1.15139924993856e-05\\
50	1.25348843777241e-05\\
51	1.43939122754167e-05\\
52	1.5310090250562e-05\\
53	1.6797589115597e-05\\
54	1.79889403651503e-05\\
55	1.98386074590579e-05\\
56	2.23623532172781e-05\\
57	2.5278021594965e-05\\
58	2.79283485872456e-05\\
59	3.07458639703615e-05\\
60	3.45416578038701e-05\\
61	3.92839780246643e-05\\
62	4.35687517928037e-05\\
63	4.94583150704574e-05\\
64	5.49924878952007e-05\\
65	5.85765645145803e-05\\
66	6.26984956068236e-05\\
67	6.76880883751089e-05\\
68	7.35202118754659e-05\\
69	8.22620918699372e-05\\
70	9.3774092789478e-05\\
71	0.000106873530309346\\
72	0.00012383738990468\\
73	0.000147348704531287\\
74	0.000165069827299666\\
75	0.000195739826276473\\
76	0.000225989189790043\\
77	0.000225989189789969\\
78	0.000225989189789943\\
79	0.000225989189790149\\
};
\addlegendentry{femur}

\addplot [color=mycolor2,dashed]
  table[row sep=crcr]{%
1	1.69254239427559e-05\\
2	9.01482658389998e-06\\
3	6.85872331753108e-06\\
4	5.51472223742167e-06\\
5	4.63306237352784e-06\\
6	3.55512173228404e-06\\
7	2.9497178083605e-06\\
8	2.40879780676307e-06\\
9	2.07434282863878e-06\\
10	1.97132021713704e-06\\
11	1.77375097228834e-06\\
12	1.5771335231267e-06\\
13	1.52851856892777e-06\\
14	1.39445064934478e-06\\
15	1.31048694827401e-06\\
16	1.26265697271614e-06\\
17	1.1984297178468e-06\\
18	1.18855501409171e-06\\
19	1.20579061353358e-06\\
20	1.17845512538718e-06\\
21	1.16395274284371e-06\\
22	1.21420956295003e-06\\
23	1.23366087020552e-06\\
24	1.26820802299537e-06\\
25	1.34800401897972e-06\\
26	1.43456461766151e-06\\
27	1.53996604585533e-06\\
28	1.66655612952933e-06\\
29	1.91685859236103e-06\\
30	2.05198759146668e-06\\
31	2.24817677419705e-06\\
32	2.42515684275033e-06\\
33	2.91163992598081e-06\\
34	3.18963617478417e-06\\
35	4.01526230359157e-06\\
36	4.80574013098715e-06\\
37	6.04583627291559e-06\\
38	7.50339253493982e-06\\
39	9.13499464779399e-06\\
40	1.05331699277616e-05\\
41	1.51510226950986e-05\\
42	1.68815907857392e-05\\
43	1.89569460652676e-05\\
44	2.25161073033527e-05\\
45	2.70384956150853e-05\\
46	2.80810976908509e-05\\
47	2.99621465781804e-05\\
48	3.17397566604396e-05\\
49	3.29081740953818e-05\\
50	3.5784518751388e-05\\
51	3.84667955751248e-05\\
52	4.22948960476534e-05\\
53	4.55162501743882e-05\\
54	4.89900397796685e-05\\
55	5.23217379286876e-05\\
56	5.85745104919229e-05\\
57	6.22581403111782e-05\\
58	6.52252064117951e-05\\
59	6.71448507002199e-05\\
60	7.27419463705747e-05\\
61	7.4168559245757e-05\\
62	7.54108529511763e-05\\
63	7.67179766966404e-05\\
64	7.84840467546313e-05\\
65	7.91911207869033e-05\\
66	7.99016631700508e-05\\
67	8.112703452255e-05\\
68	8.18970485870859e-05\\
69	8.26862983314275e-05\\
70	8.38429209130684e-05\\
71	8.48614376360054e-05\\
72	8.56348979109117e-05\\
73	8.70332780624865e-05\\
74	8.82970112549361e-05\\
75	8.94310036124727e-05\\
76	9.06690743290199e-05\\
77	9.2156178401853e-05\\
78	9.31953100842492e-05\\
79	9.84602534001091e-05\\
80	0.000100698295811563\\
81	0.000106681154295734\\
82	0.000110704341036683\\
83	0.0001192330176178\\
84	0.000126157255458801\\
85	0.000126157255466119\\
86	0.000126157255462031\\
87	0.000126157255464289\\
};
\addlegendentry{lung}

\addplot [color=black,dash dot]
  table[row sep=crcr]{%
1	4.59663916700538e-05\\
2	2.7634710972469e-05\\
3	1.75937828511836e-05\\
4	1.59713563564706e-05\\
5	1.32497678907995e-05\\
6	1.0976519056421e-05\\
7	1.03943681506129e-05\\
8	1.00198022678329e-05\\
9	1.01060616226445e-05\\
10	1.12407459575546e-05\\
11	1.1890198149157e-05\\
12	1.22412864872956e-05\\
13	1.34565238852681e-05\\
14	1.35519935836887e-05\\
15	1.37523985086444e-05\\
16	1.42133101816992e-05\\
17	1.47121968725357e-05\\
18	1.49769701615816e-05\\
19	1.53885962735295e-05\\
20	1.62706933483251e-05\\
21	1.71081721194561e-05\\
22	1.79757684853038e-05\\
23	1.89597023338942e-05\\
24	2.00861431032606e-05\\
25	2.10215194411937e-05\\
26	2.15521169041569e-05\\
27	2.20719169379612e-05\\
28	2.29195622647551e-05\\
29	2.45556278284233e-05\\
30	2.55659879263028e-05\\
31	2.604109761708e-05\\
32	2.64541697928836e-05\\
33	2.71447085177426e-05\\
34	2.759770692604e-05\\
35	2.8292421900149e-05\\
36	2.9132564817884e-05\\
37	3.00680952497749e-05\\
38	3.07280580300509e-05\\
39	nan\\
};
\addlegendentry{hand}

\addplot [color=mycolor3, draw=none, mark=square*, mark options={solid, mycolor3}]
  table[row sep=crcr]{%
15	3.75671986166275e-06\\
10	1.97132021713704e-06\\
9	1.01060616226445e-05\\
};
\addlegendentry{proposed}

\addplot [color=mycolor4, draw=none, mark=*, mark options={solid, mycolor4}]
  table[row sep=crcr]{%
52	1.5310090250562e-05\\
87	0.000126157255464289\\
15	1.37523985086444e-05\\
};
\addlegendentry{white noise \cite{Nadakuditi08}}

\addplot [color=mycolor5, draw=none, mark=triangle*, mark options={solid, mycolor5}]
  table[row sep=crcr]{%
13	4.69080179778324e-06\\
8	2.40879780676307e-06\\
5	1.32497678907995e-05\\
};
\addlegendentry{95\% \cite{Lindner13}}

\addplot [color=mycolor6, draw=none, mark=diamond*, mark options={solid, mycolor6}]
  table[row sep=crcr]{%
5	1.53691460926249e-05\\
7	2.9497178083605e-06\\
4	1.59713563564706e-05\\
};
\addlegendentry{nonuniform \cite{Aouada04}}

\addplot [color=Mei, draw=none, mark=star, mark options={solid, Mei}]
  table[row sep=crcr]{%
9	7.95631136597516e-06\\
2	9.01482658389998e-06\\
2	2.7634710972469e-05\\
};
\addlegendentry{t-test \cite{Mei08}}

\end{axis}
\end{tikzpicture}%
		\caption{$e_{\text{LMMSE}}$ for all possible order $t$. The evaluated model orders are obtained using all available samples to train the PDMs. The order selected by ``white noise" (orange circle) for the lung data set is not depicted because is out of plot limits. 
} 
		\label{fig:LMMSE}
\end{figure}
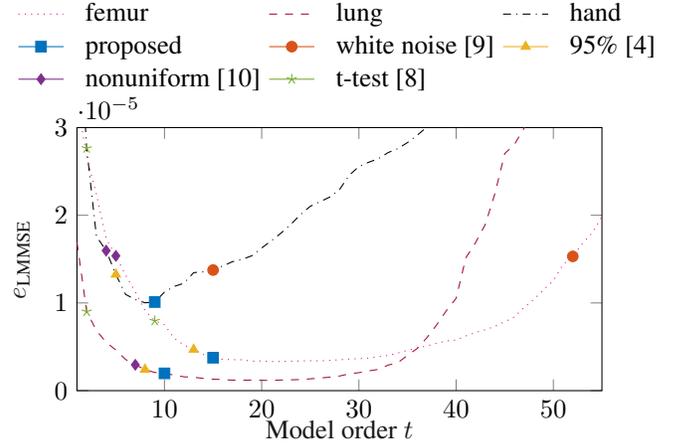

As an illustrative example of the importance of the model order in statistical shape models, we performed an experiment that shows how well the PDM with the selected model order can deal with partial occlusions. Considering an obtained PDM as the prior information about shape deformation, we perform an estimate of an occluded (or missing) landmark in a new observed shape. The test consists in the following: inside a leave-one-out test, this is, for all $m = 1, \dots, M$ available samples in a data set, within the $m$th ``left-out" sample we delete one landmark from that shape. Then, we estimate it from the remaining landmarks using a linear minimum mean-squared error (LMMSE) estimator. The $M-1$ ``not left-out" samples are used to design the PDM, in which we evaluate all possible orders $t$. 
Let ${\bf y}_i$ denote the missing landmark (which consists of its $x$- and $y$-coordinates) and ${\bf y}_a$ the remaining available landmarks. Let $\hat{\mathbf{R}}_{ia}$ denote the sample cross-covariance matrix between the missing landmark and the remaining available landmarks, and $\hat{\mathbf{R}}_{aa}$ the covariance matrix of the available landmarks. These matrices are calculated from the available PDM of order $t$. The LMMSE estimator of the missing landmark from the remaining landmarks is then $\hat{\mathbf{y}}_i = \hat{\mathbf{R}}_{ia}\hat{\mathbf{R}}_{aa}^{-1}\mathbf{y}_a$. In our experiment, we successively estimate one landmark $i$ from the others, repeating this for all $i = 1, ..., N/2$ landmarks. We average the error over all landmarks and over all available samples and obtain 
\begin{equation}
e_{\text{LMMSE}}(t)=\frac{1}{M}\frac{1}{N/2}\sum_{m=1}^{M}\sum_{i=1}^{N/2}||\hat{\mathbf{y}}^{(m)}_i-\mathbf{y}^{(m)}_i||^2,
\end{equation}
which is evaluated for all possible model orders, i.e., $t =1, \ldots, \text{min}(N,M)$. Figure \ref{fig:LMMSE} shows this metric for the three data sets as a function of considered model order $t$. 
In Fig. \ref{fig:LMMSE}, we observe that the evaluated empirical error decreases until it reaches a minimum, which is different for each data set. A shape model with too small an order may suffer from underfitting (thus, the error decreases if we add more complexity), and too large an order may lead to overfitting (and therefore, the error decreases if we reduce the complexity). 
We conclude that choosing the right model order is critical in order to minimize the LMMSE. We see that the model order determined by our technique (blue squares) leads to the smallest LMMSE among all competing techniques. We also notice that there is a relatively large interval of model orders that lead to similar LMMSEs. The principle of parsimony dictates that in such a case a smaller order is to be preferred. Our technique observes this principle. 

\section{Conclusion}
Statistical shape models provide important information to machine learning algorithms about object deformation. The order of these models is typically obtained heuristically. We have proposed a model-order selection strategy that is based on information-theoretic criteria and thus has a theoretical justification. 
We have validated the selection performance of our technique on simulated shape data, and it outperformed other model-order selection strategies under different conditions of sample support and noise level. We have also evaluated the technique on shapes from real data sets, showing results similar to the evaluation with artificial data. Additionally, we have performed an empirical test to illustrate the impact of the model order of shape models, and how the choice of order provided by our technique results on a model with better performance.

\section{Acknowledgement}
The authors would like to thank Prof. Barry Quinn for insightful discussions. 

\bibliographystyle{IEEEbib}
\bibliography{references}

\end{document}